\documentclass[runningheads]{llncs}
\usepackage{graphicx}
\usepackage{comment}
\usepackage[utf8]{inputenc}
\usepackage{amsmath,amssymb} 
\usepackage{color}

\usepackage{xspace}

\makeatletter
\DeclareRobustCommand\onedot{\futurelet\@let@token\@onedot}
\def\@onedot{\ifx\@let@token.\else.\null\fi\xspace}

\def\eg{\emph{e.g}\onedot}

\def\etal{\emph{et al}\onedot}
\makeatother

\usepackage{graphicx}
\graphicspath{ {./imgs/} }
\usepackage{subfig}

\begin{document}
\pagestyle{headings}
\mainmatter

\title{MaskedFusion: Mask-based 6D Object Pose Estimation} 

\titlerunning{MaskedFusion}
%
\author{Nuno Pereira \orcidID{0000-0001-7177-751X} \and
Luís A. Alexandre \orcidID{0000-0002-5133-5025} }
\authorrunning{N. Pereira and L.A. Alexandre}
%
\institute{NOVA LINCS, Universidade da Beira Interior, Covilhã, Portugal\\
\email{\{nuno.pereira,luis.alexandre\}@ubi.pt}}
\maketitle

\begin{abstract}
    MaskedFusion is a framework to estimate the 6D pose of objects using RGB-D data, with an architecture that leverages multiple sub-tasks in a pipeline to achieve accurate 6D poses.
    6D pose estimation is an open challenge due to complex world objects and many possible problems when capturing data from the real world, \eg, occlusions, truncations, and noise in the data.
    Achieving accurate 6D poses will improve results in other open problems like robot grasping or positioning objects in augmented reality.
    MaskedFusion improves the state-of-the-art by using object masks to eliminate non-relevant data.
    With the inclusion of the masks on the neural network that estimates the 6D pose of an object we also have features that represent the object shape.
    MaskedFusion is a modular pipeline where each sub-task can have different methods that achieve the objective.
    MaskedFusion achieved $97.3\%$ on average using the ADD metric on the LineMOD dataset and $93.3\%$ using the ADD-S AUC metric on YCB-Video Dataset, which is an improvement, compared to the state-of-the-art methods. The code is available on GitHub (\url{https://github.com/kroglice/MaskedFusion}).

\end{abstract}

\section{Introduction}
\begin{figure}[thpb]
   \centering
   \includegraphics[width=\textwidth]{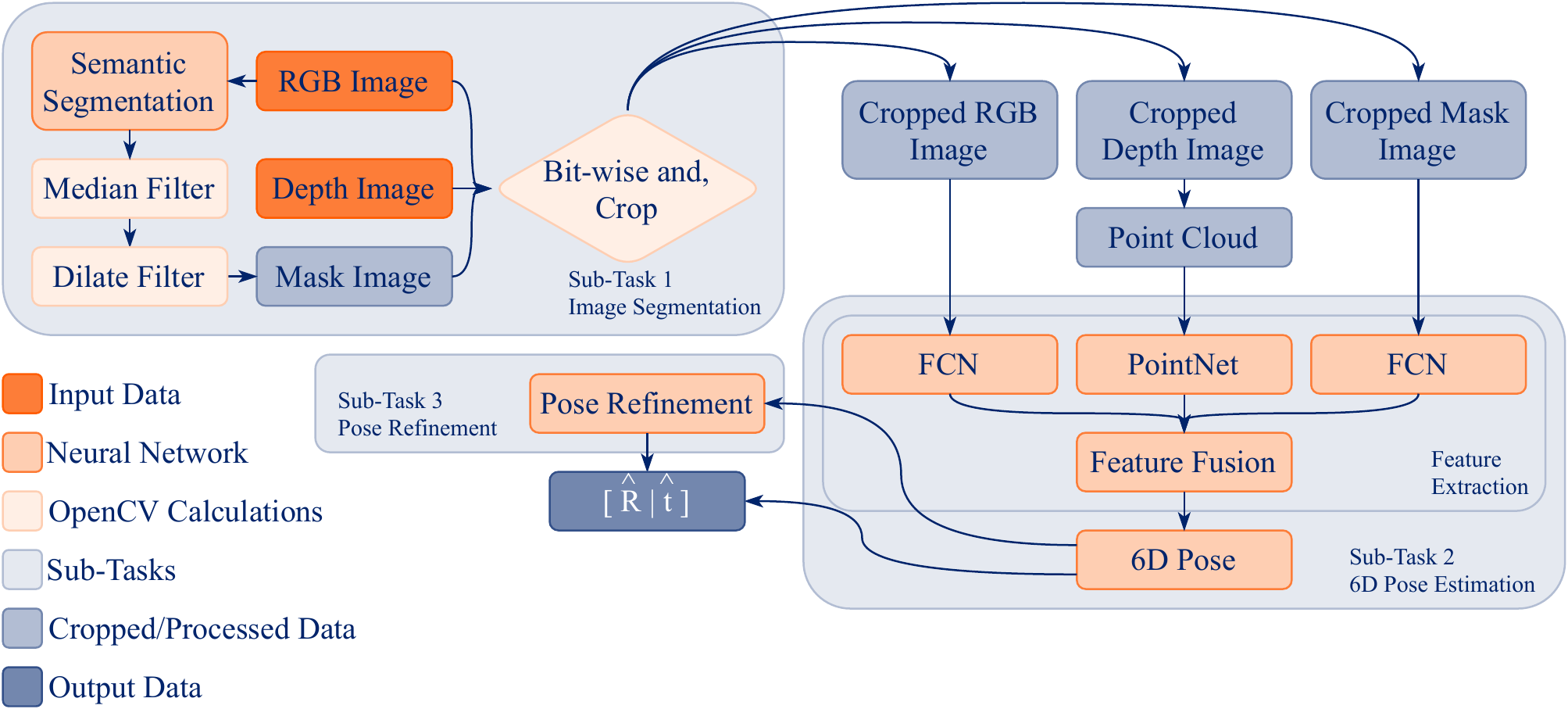}
   \caption{Pipeline diagram of the MaskedFusion architecture. MaskedFusion has three sub-tasks: image segmentation, 6D pose estimation, and a pose refinement}
   \label{fig:digarch}
\end{figure}
Object detection and pose estimation are important tasks for robotic manipulation, scene understanding and augmented reality, and it has recently been the target of much research effort.
With the increasing automation and the need for robots that can work in non-restricted environments, the capacity to understand the scene in 3 dimensions is becoming a must.
One of the main tasks involving robots is grasping objects.
Performing gasping in a non-restricted or/and cluttered environment, \eg, bin picking, is a hard problem to tackle.
6D pose estimation is a task in computer vision that detects the 6D pose (3 degrees of freedom for the position and the other 3 for orientation) of an object.
6D pose estimation is an open important problem because it can be used in several important tasks like grasping, robotic manipulation, augmented reality and others.
6D pose is as important in robotic tasks as in augmented reality, where the pose of real objects can affect the interpretation of the scene and the pose of virtual objects can also improve the augmented reality experience.
It can also be useful in human-robot interaction tasks such as learning from demonstration and human-robot collaboration.
Estimating the object's 6D pose is a challenging problem due to the different objects that exist and how they appear in the real world.
Captured scenes from the real world might have occlusions and truncations on some objects.
Obtaining the data to retrieve the 6D pose is also a problem, as RGB-D data can be hard to obtain for certain types of object, \eg, fully metallic objects and meshed office garbage bins.
6D pose estimation can be split into three different categories, defined by the type of input data that the methods use.
Methods \cite{pnpex1}, \cite{deepex2}, \cite{pnpex2}, \cite{pvnet}, \cite{deepex1}, \cite{pnpex3}, \cite{pnpex4} that use RGB images as input usually rely on the detection and matching of keypoints from the objects in a scene with the 3D render and use the \textit{PnP} \cite{pnp} algorithm to solve the pose of that object.
Point cloud methods \cite{pvfh},  \cite{frustum}, \cite{pointnet}, \cite{fpfh}, \cite{vfh}, \cite{voxelnet} rely on descriptors to extract features from the objects in scene to later be matched with features captured in known poses.
Finally, RGB-D methods \cite{ssd6d}, \cite{localrgbd}, \cite{liunified}, \cite{densefusion}, \cite{posecnn}, \cite{pointfusion} have the 6D poses directly regressed from the data, and then furthered refined.
Most of the methods in this category only use the depth data in the refinement phase.
Our method, MaskedFusion, fits in the RGB-D category because we directly regress the 6D pose from the RGB-D data.

We propose a new method based on a pipeline with 3 sub-tasks that, when combined, can estimate the object's 6D pose.
For the first sub-task, we detect and classify each object in the scene using semantic segmentation and retrieve their binary masks.
Then, in the second sub-task, for each object detected, we extract features from the different types of data and fuse them in a pixel-wise manner.
After the fusion, we have the 6D pose neural network (NN) that will regress the 6D pose of the object.
The third sub-task is optional but advisable and refines the 6D pose of the object.
For the refinement sub-task, we used a NN that was proposed in \cite{densefusion}.

MaskedFusion improves upon the state-of-the-art method by achieving greater scores in two (LineMOD \cite{linemod}, YCB-Video \cite{posecnn}) of the most used datasets in the 6D pose estimation area.
Our method achieved an average of $97.3\%$ and our best experience achieved $97.8\%$ in the LineMOD dataset using the ADD (\ref{eq:add}) metric.
For the YCB-Video dataset, we achieved on average $93.3\%$ using the ADD-S (\ref{eq:add-s}) AUC metric and $97.1\%$ of ADD-S smaller than $2cm$.

In summary, our work has the following main contributions:
\vspace{-.7em}
\begin{itemize}
    \item End-to-end modular pipeline to estimate the object's 6D pose;
    \item A new neural network to estimate 6D pose that can be used outside of our pipeline, it is also very fast to execute during inference;
    \item Improvements to the state-of-the-art on the two most used 6D pose evaluation datasets.
\end{itemize}


\section{Related Work}
In this section, we present the most relevant literature.
In subsection \ref{sub:ss} we present methods of semantic segmentation that use CNNs and FCNs.
On the following subsection \ref{sub:6d} we present the relevant literature for 6D pose estimation.


\subsection{Semantic Segmentation}
\label{sub:ss}
The semantic segmentation goal is to classify each pixel in an image.
In contrast, the instance segmentation goal is to detect, delineate it with a bounding box and segment or create a mask for each object instance present in the image.
The objective of the panoptic segmentation task \cite{panopticsegmentation} is the unification of the two typically distinct tasks of semantic segmentation and instance segmentation for a given image.


Some of the most notable techniques used in semantic segmentation are presented in this section starting with the U-Net \cite{unet}.
It is most known for its U-shaped architecture that does semantic segmentation with the encoder-decoder method.
The authors of U-net created its architecture with two consecutive convolutions followed by a max-pooling.
With this method, the spatial information is decreased but feature information is increased.


U-Net and SegNet \cite{segnet} have similar architectures where the second half of those architectures consists of the same structure in the first half but hierarchically opposite.
SegNet is a FCN based on the 13 VGG-16 \cite{VGG16} convolutional layers.
Another technique commonly found in semantic segmentation is addressing it using contextual information.


Pyramid Scene Parsing Network (PSPNet) \cite{pspnet}, also uses global contextual information for semantic segmentation.
The authors introduced a Pyramid Pooling Module after the last layers of an FCN (based on ResNet-18), the feature maps obtained from the FCN are pooled using 4 different scales corresponding to 4 different pyramid levels each with different sizes, $1\times1$, $2\times2$, $3\times3$ and $6\times6$.
The polled feature maps are then convoluted using a $1\times1$ convolution layer to reduce the feature maps dimension.
So the outputs of each convolution are up-sampled and then concatenated with the initial feature maps that were obtained in the FCN.
This concatenation provides the local and global contextual information of the image.
After the concatenation, the authors use another convolution layer to generate the final pixel-wise predictions.
The main idea of PSPNet is to observe the whole feature map in sub-regions with different locations using the pyramid pooling module, such that, the network achieves a better understanding of the scene.

\subsection{6D Pose Estimation}
\label{sub:6d}
Methods that estimate object 6D pose can be split into three different categories, RGB, Point Cloud, RGB-D.
These categories are defined by the type of input data that methods use.

Methods that use RGB images as input \cite{pnpex1}, \cite{deepex2}, \cite{pnpex2}, \cite{pvnet}, \cite{deepex1}, \cite{pnpex3}, \cite{pnpex4} usually rely on the detection and matching of keypoints from the objects in a scene with the 3D render and use the \textit{PnP} \cite{pnp} algorithm to solve the pose of that object.
Although they are very fast methods, their recall drops quickly in the presence of occlusion.
The methods that do not use this technique are methods that use deep learning \cite{deepex2}, \cite{deepex1}.
These methods rely on convolutional neural networks (CNNs) to extract and store features from the multiple viewpoint.
This process is done during the training phase of the CNN.
On the inference phase, features are extracted from the new scenes and then matched with previously known features and with this matching process it is possible to obtain the object 6D pose.

Wu \etal \cite{wu2018real} method is made to be fast, running at 20Hz on RGB images with accurate 6D object poses achieved from segmenting the RGB image to obtain the masks and the classes of the objects and then for each mask, with the use of a pose interpreter network, it regresses the 6D pose of the object.
The pose interpreter network learns with a new proposed loss similar to $L_1$ but applied on the point clouds generated from the objects.
One of the most accurate method in 6D pose using RGB images is PVNet \cite{pvnet}.
PVNet is a hybrid method that uses the two methods, segmentation and then \textit{PnP}. 
With this combination, it can handle mild occlusion and truncations.
PVNet has a NN to predict pixel-wise object labels and unit vectors that represent the direction from each pixel to every keypoint of the object. These keypoints are voted and matched with the 3D model of the object to estimate its 6D pose.

Point cloud methods \cite{pvfh}, \cite{frustum}, \cite{pointnet}, \cite{fpfh}, \cite{vfh}, \cite{voxelnet} rely on descriptors to extract features from the objects in the scene to later be matched with features captured in known poses.
Methods like PVFH \cite{pvfh} and its predecessors \cite{fpfh}, \cite{vfh} achieve remarkable speed acquiring the 6D pose of the object, but these methods need good data retrieval to be accurate and most of the data captured from the real world usually has different types of interference and/or noise.
So deep learning is also contributing in this category of methods to achieve better accuracy and get better extracted features from the objects, mitigating some of the noise and interference that can appear.
Many deep learning architectures were propose, like PointNets \cite{frustum}, \cite{pointnet} and VoxelNet \cite{voxelnet}. These methods achieved good results in multiple datasets.

Finally, RGB-D methods \cite{ssd6d}, \cite{localrgbd}, \cite{liunified}, \cite{densefusion}, \cite{posecnn}, \cite{pointfusion} usually have the object 6D pose directly regressed from the data, and usually further refined by other methods, \eg, Iterative Closest Point (ICP).
Tejani \etal \cite{tejani} follow a local approach where small RGB-D patches vote for object pose hypotheses in a 6D space.
Kehl \etal \cite{localrgbd} also follow a local approach but they use a convolutional auto-encoder (CAE) to encode each patch of the object to later match these features obtained in the bottleneck of the CAE with a code-book of features learned during the train, and use the code-book matches to predict the 6D pose of the object.
Although such methods are not taking global context into account, they proved to be robust to occlusion and the presence of noise artifacts since they infer the object pose using only small patches of the image.
SSD-6D \cite{ssd6d} uses an RGB image that is processed by the NN to output localized 2D detection with bounding boxes and then it classifies the bounding boxes into discrete bins of Euler angles and subsequently estimates the object 6D pose.
This method is in the RGB-D category because after the first estimation, and with the availability of the depth information, the 6D poses can be further refined.
PoseCNN \cite{posecnn} uses a new loss function that is robust to object symmetry to directly regress the object rotation.
It uses a Hough voting approach to obtain the 3D location center of the object to achieve its translation.
Using ICP on the refinement phase of SSD-6D and PoseCNN makes their 6D pose estimation accurate.
Li \etal \cite{liunified} formulates a discriminative representation of 6-D pose that enables predictions of both rotation and translation by a single forward pass of a CNN, and it can be used with many object categories.
DenseFusion \cite{densefusion} extract features from RGB images and depth data with different FCN.
After the extraction, it fuses the depth feature with the RGB features while retaining the input's space geometric structure, and then it estimates the object 6D pose. 
DenseFusion is similar to PointFusion \cite{pointfusion}, as it also estimates the 6D pose while keeping the geometric structure and appearance information of the object, to later fuse this information in a heterogeneous architecture.

\section{Methodology}
Our method, MaskedFusion, is a pipeline (Fig. \ref{fig:digarch}) dived into 3 sub-tasks that combined can solve the task of object 6D pose estimation.
MaskedFusion is a modular pipeline, for each sub-task we use a NN to solve the task that it is responsible for.
However, since our pipeline is modular every sub-task can have different types of methods that will solve the task in hand and they can be replaced easily.

In the first sub-task, is where we need to detect and segment each object in the scene. To do that we use a NN based on the encoder-decoder architecture to classify each pixel of the RGB image captured, and predict the mask and the location for each object in the scene.

For the second sub-task, with the masks obtained in sub-task 1 for each object and the RGB-D data, we can now estimate the object 6D pose.
After this sub-task, we obtain the rotation matrix and the translation vector according to the 6D pose estimated from the 6D pose NN.
After obtaining the translation vector and rotation matrix we can use another method to refine further the estimated 6D pose.
This last sub-task is optional but advisable for better results.

In the last sub-task, we use the same refinement network as the DenseFusion \cite{densefusion}.
This last step can be slow during the training because it needs many training epochs to show its advantages, but during the inference, it is very fast.
Other methods usually use ICP that is also a good method for refinement but it has a higher computational cost and will take longer to apply the refinements during the inference time.

During the training phase of our pipeline, both of the main NNs, semantic segmentation and 6D pose, are trained independently.

\subsection{Semantic Segmentation}
Semantic labeling provides richer information about the objects and handles occlusions better than object detection and localization methods.
For the first sub-task of our pipeline, a semantic segmentation method is used to detect and extract the mask of each object presented in the scene.
For the completion of this sub-task, we only use the RGB image.

For the semantic segmentation, we use an FCN with encoder-decoder architecture.
Our implementation is similar to Segnet \cite{segnet}, however, since our pipeline is modular it is possible to use other methods to detect and obtain the object masks.
We used and tested with a semantic segmentation method but it is also possible to use an instance segmentation, panoptic segmentation, or any other methods that can detect and output the mask for each object in the scene.

In our implementation of the semantic segmentation, after the output of the mask, we apply 2 filters on the mask before saving or feed-forward it to the next sub-task.
First, we use the median filter from \textit{OpenCV} to smooth the image with a kernel size of $3\times3$.
Second, we dilate the mask with a $5\times5$ kernel such that, if the mask has some minor boundary segmentation error, this operation helps to correct it.

The binary mask obtained from the semantic segmentation method with the 2 filters applied is used to crop the RGB and depth images. This is done for each object present in the scene.
To crop the RGB and depth images, we apply a \textit{bit-wise and} between the RGB image and mask, and also between the depth image and the mask.
The result of the \textit{bit-wise and} of the RGB image and the mask will be inside of a rectangular crop that encloses all the object and this smaller image will serve as input data to the 6D pose NN.
In the case of the depth image, a point cloud is further generated from the resultant \textit{bit-wise and} cropped depth image, and the point cloud will also serve as input to the next phase of our pipeline.
Cropping the data with the mask is a pre-processing of the data that helps the 6D pose NN on the second sub-task because it discards the background or other non-relevant data that are around the object leaving only the data that is most relevant to the 6D pose estimation.

\subsection{6D Pose Neural Network}
Our NN that estimates the 6D pose of the object is inspired in DenseFusion \cite{densefusion} and PointFusion \cite{pointfusion}.
It has a similar architecture that is used to extract features from different types of data and fuse the information in a pixel-wise manner.
Our method improves upon DenseFusion and others, and we show our performance in the results section (\ref{sec:results}).
Fig. \ref{fig:digarch} shows the architecture of our 6D pose NN (Sub-task 2).
Our 6D pose NN can be divided into two stages: feature extraction and 6D pose estimation.

On the first stage, feature extraction, all the data that were cropped after the semantic segmentation are separated into different NNs that extract features for each type of data.
For the point cloud data that were generated from the cropped depth image, we used the PointNet \cite{pointnet} to extract 500 features that represent the depth information of the object.
For the cropped RGB image we use a fully convolutional neural network (FCN) based on \textit{ResNet-18} to also extract 500 features from the color image.
Our \textit{ResNet-18} is similar to the original but without the fully connected layers at the end making it an FCN.
Finally, for the binary mask image, we also used an FCN similar to the FCN used for the color images, but in this case, at the start, we only have one channel as input instead of three.
As in the color images for the masks, we extract 500 features.
Having features from the mask enables us to have features that represent the shape of the object which is more relevant information for our next stage and improved the accuracy of our method.

On the second stage, all the extracted features of each data source are then concatenated into a single vector and pass through a convolutional layer to fuse all the features.
We then use another NN that receives the features extracted previously and regresses the 6D pose of the object, that is, the rotation matrix and the translation vector of the object.
With all the features concatenated from each input, and then feed-forward to a convolutional layer, we will have 2 different paths (each with 4 convolutional layers), one of the paths is for the regression of the translation vector and the other is to regress the rotation matrix.

So that our NN can regress the 6D pose we use (\ref{eq:loss}) as loss function during the train.
This loss function is the same that the DenseFusion used:
\vspace{-.2em}
\begin{equation}
    \label{eq:loss}
    \mathcal{L}^{p}_{i} = \frac{1}{M} \sum_{j} \left \| ( Rx_j + t) - (\hat{R_i}x_j + \hat{t_i}) \right \|
\vspace{-.6em}
\end{equation}

where, $x_j$ denotes the $j^{th}$ point of the $M$ randomly selected 3D points from the objects 3D model, $p = [R|t]$ is the ground truth pose, where $R$ is the rotation matrix of the object and $t$ is the translation.
The predicted pose generated from the fused embedding of the $i^{th}$ dense-pixel is represented by $\hat{p}_i = [ \hat{R}_i| \hat{t}_i]$ where $\hat{R}$ denotes the predicted rotation and $\hat{t}$ the predicted translation.
After training the 6D pose NN, the output of it ($\hat{p}_i = [ \hat{R}_i| \hat{t}_i]$) can be retrieved after the second stage or it can be sent to the next sub-task of the pipeline.

\subsection{Pose Refinement}
In the last sub-task, we used the same pose refinement that DenseFusion developed.
The authors of DenseFusion created a pose refinement NN that improves upon the 6D pose previously estimated.
The output 6D poses estimated before serves as input to the DenseFusion pose refinement NN.

We tested the DenseFusion refinement NN and concluded that it is very slow during the training because it needs many training epochs to start showing its advantages, but during the inference, it is very fast.
Other refinement methods, can be used but they usually require more computation during the inference and it slows down the process of estimating the 6D pose.

\section{Experiments}
We tested our method on two widely-used datasets, LineMOD \cite{linemod} and YCB-Video \cite{posecnn}.
We use these datasets because it is easier to compare our method with previous methods that were also tested in the same datasets.

\begin{figure}[thpb]
\vspace{-2em}
    \centering
    \subfloat[LineMOD]{{\includegraphics[width=.3\linewidth]{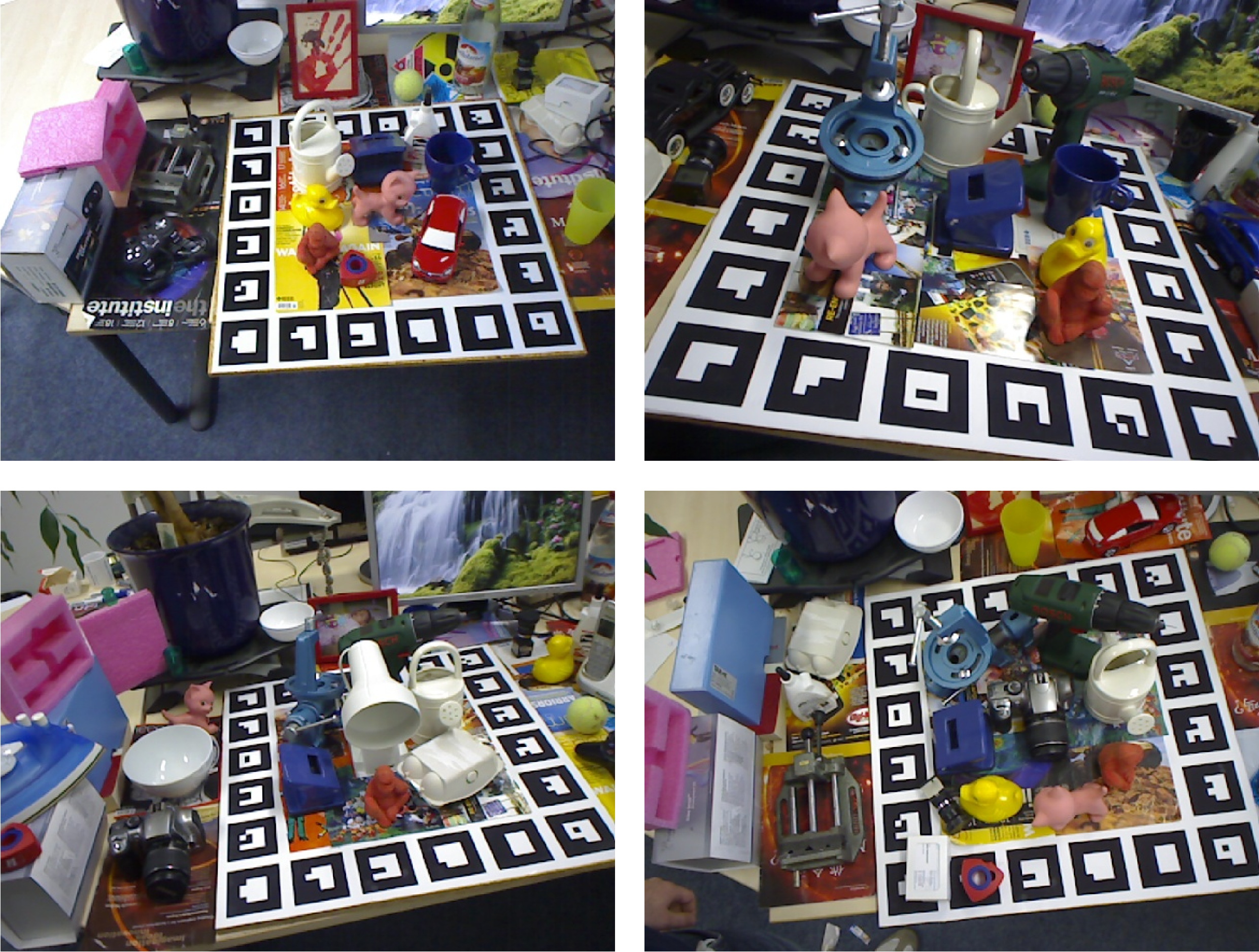} }}%
    \qquad
    \subfloat[YCB-Video]{{\includegraphics[width=.3\linewidth]{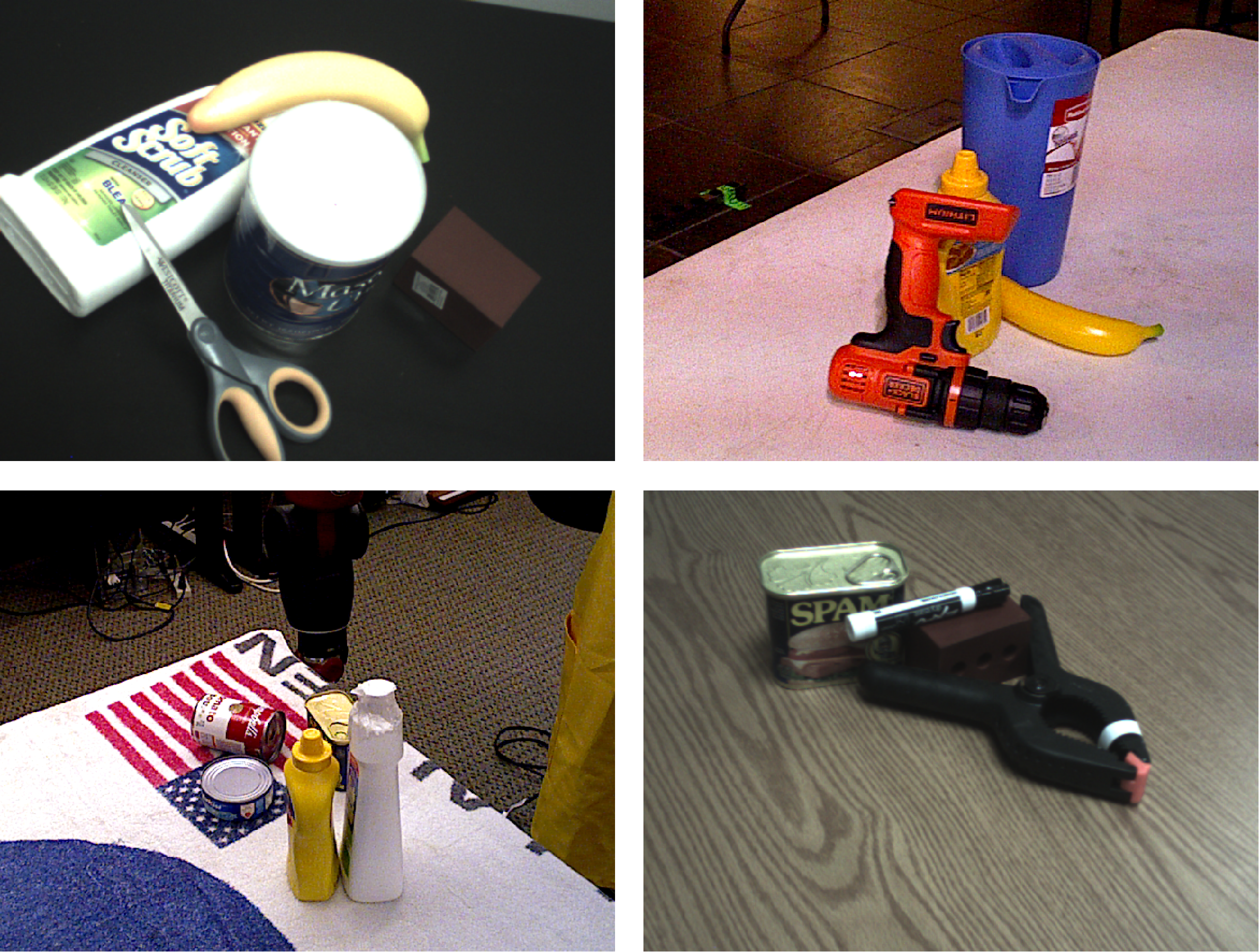} }}%
    \caption{Example of images from the datasets}
    \label{fig:linemod}%
\vspace{-2em}
\end{figure}
For each experiment we trained the methods from scratch and then evaluate them.
We repeat this procedure 5 times and present the average results, whereas other references usually only present one result.
All experiments were executed on a desktop with SSD NVME, 64GB of RAM, an NVIDIA GeForce GTX 1080 Ti and Intel Core i7-7700K CPU.

\subsection{LineMOD}
LineMOD \cite{linemod} is one of the most used datasets to tackle the 6D pose estimation problem.
Many types of methods that tackle the 6D pose estimation problem use this dataset ranging from the classical methods like \cite{lineex1}, \cite{lineex2}, \cite{lineex3} to the most recent deep learning approaches \cite{lineex4}, \cite{densefusion}, \cite{posecnn}.
This dataset was captured with a Kinect, and it has a procedure that automatically aligns the RGB and depth images.
LineMOD has 15 low-textured objects (although we only use 13 as in previous methods) in over 18000 real images and has the ground truth pose annotated.

Each object is associated with a test image showing one annotated object instance with significant clutter but only mild occlusion, as shown in Fig. \ref{fig:linemod}.
Each object also contains the 3D model saved as a point cloud and a distance file with the maximum diameter ($cm$) of the object.

\subsection{YCB-Video}
After the release of the YCB-Video dataset in \cite{posecnn} it started being used by methods that can estimate the 6D pose of objects even in scenes with heavy occlusions.
This dataset has 21 objects that were presented in \cite{originalycb} by Calli \etal. 
These objects have different shapes and textures, and mild occlusions.
YCB-Video is composed of 92 RGB-D videos, each with a subset of the objects placed in the scene.
These videos have the segmentation masks and the 6D poses of the objects annotated, and for each frame there are at least 3 objects present in the scene.
We have used the dataset in the same splits has previous works, \cite{posecnn}, \cite{densefusion}, where 80 videos were used for training and 12 for testing.
The \textit{80,000} synthetic images released in \cite{posecnn} were also used during the train of our 6D pose sub-task, but not used for the semantic segmentation sub-task.

\subsection{Metrics}
As in previous works \cite{ssd6d}, \cite{pvnet}, \cite{densefusion}, \cite{posecnn} we used the Average Distance of Model Points (ADD) \cite{linemod} as metric of evaluation for non-symmetric objects and for the egg-box and glue we used the Average Closest Point Distance (ADD-S) \cite{posecnn}.
We needed to use another metric for these two objects because they are symmetric objects.
\vspace{-0.2em}
\begin{equation}
    \label{eq:add}
    \textnormal{ADD} = \frac{1}{m} \sum_{x \in M} \left \| ( Rx + t) - (\hat{R}x + \hat{t}) \right \|
\vspace{-.6em}
\end{equation}

In the ADD metric (equation \ref{eq:add}), assuming the ground truth rotation $R$ and translation $t$ and the estimated rotation $\tilde{R}$ and translation $\tilde{t}$, the average distance calculates the mean of the pairwise distances between the 3D model points of the ground truth pose and the estimated pose. In equation (\ref{eq:add}) and (\ref{eq:add-s}) $M$ represents the set of 3D model points and $m$ is the number of points.

For the symmetric objects (egg-box and glue), the matching between points is ambiguous for some poses. In these cases we used the ADD-S metric:
\vspace{-0.2em}
\begin{equation}
    \label{eq:add-s}
    \textnormal{ADD-S} = \frac{1}{m} \sum_{x_1 \in M} \min_{x_2 \in M} \left \| ( Rx_1 + t) - (\hat{R}x_2 + \hat{t}) \right \|
\vspace{-.6em}
\end{equation}

For the YCB-Video we also used the same metrics as in previous works like PoseCNN \cite{posecnn} and DenseFusion \cite{densefusion}.
We use the area under the ADD-S (\ref{eq:add-s}) curve (AUC) as in DenseFusion \cite{densefusion} where their threshold was $10cm$ the same as PoseCNN \cite{posecnn}, and we also calculate the percentage of ADD-S smaller than $2cm$, which DenseFusion \cite{densefusion} authors consider as the minimum tolerance for most of the robot grippers.
Using these same metrics in both datasets as previous works enable us to make direct comparisons with their methods.

\section{Results}
\label{sec:results}
In this section, we present the results of the experiments made and compare them with other methods that have been evaluated in the same datasets.
The inference results in subsection \ref{sec:inference} present the results for the full pipeline, meaning that we do not use the masks provided in the YCB-Video dataset, but instead rely on the masks produced by the segmentation sub-task of MaskedFusion.
This makes the task harder since the masks provided with YCB-Video are all correct whereas the ones we receive from the segmentation can have errors.

\subsection{Results in LineMOD}
We present here the results of the NN that estimates the object 6D pose of the MaskedFusion pipeline using the LineMOD dataset.

\begin{figure}[thpb]
\vspace{-2em}
    \centering
    \subfloat[Error per epoch]{{\includegraphics[width=.45\linewidth]{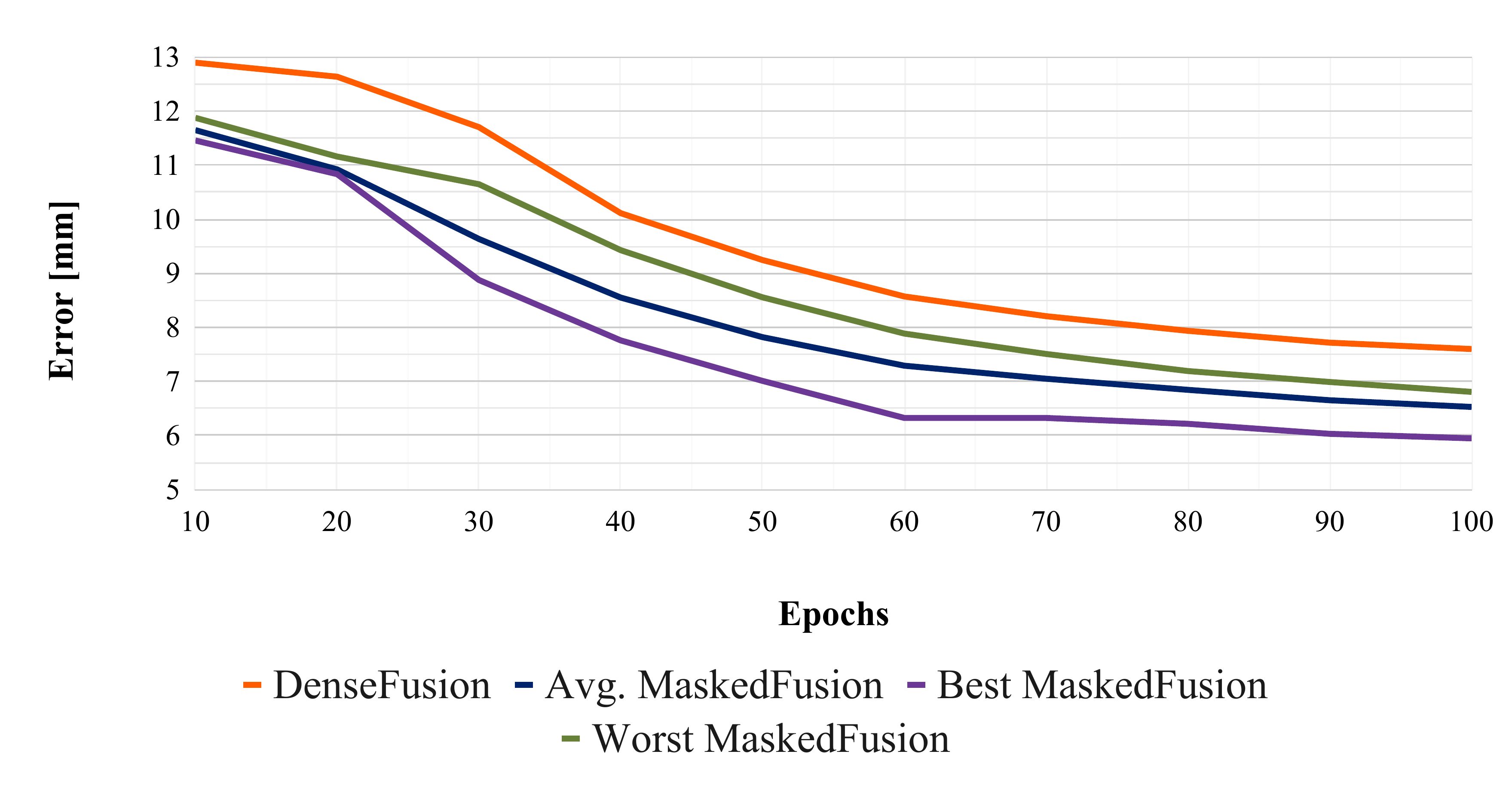} }}%
    \qquad
    \subfloat[Error per hour]{{\includegraphics[width=.45\linewidth]{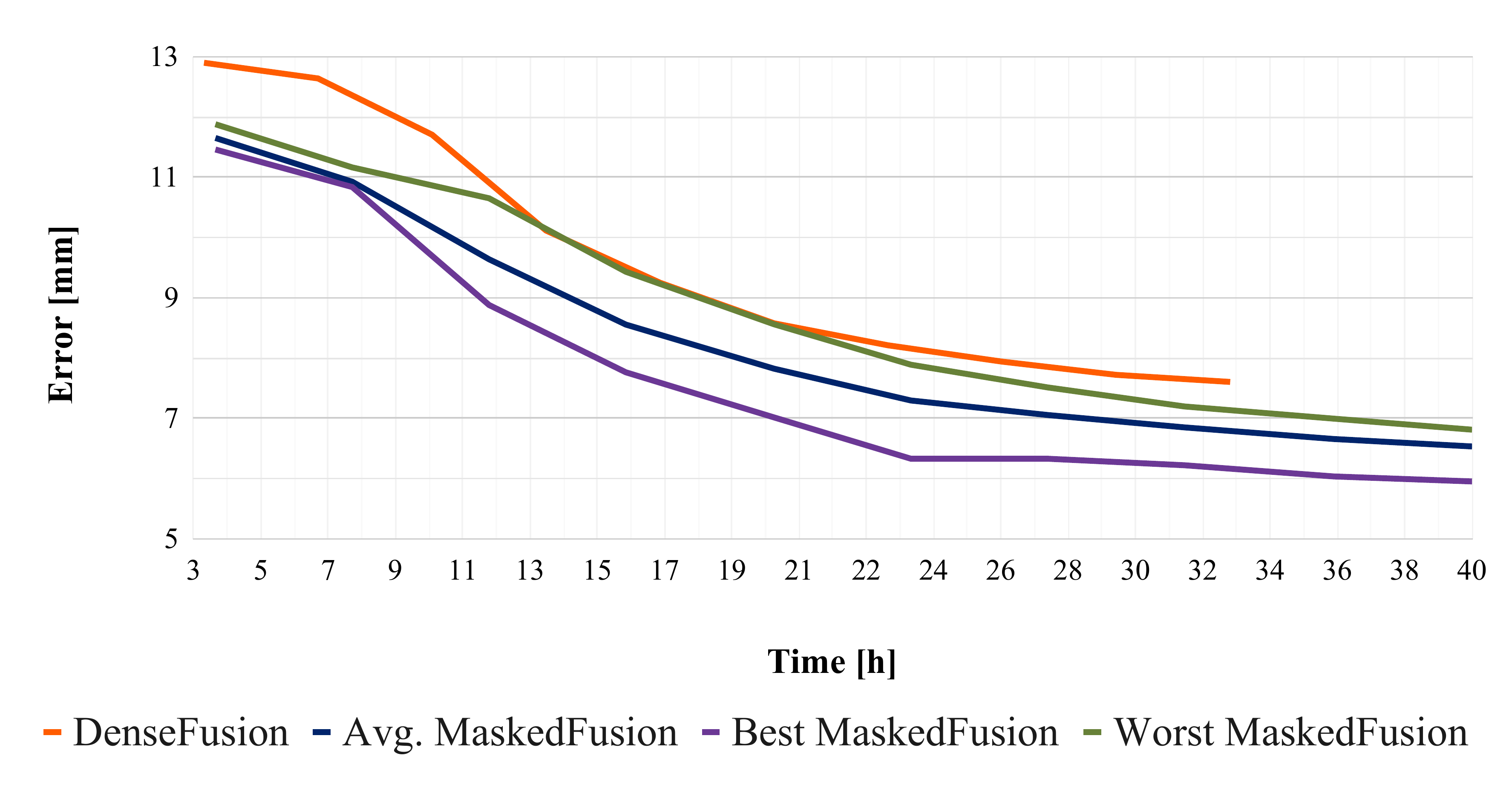} }}%
    \caption{The methods were evaluated on the test set, after every 10 training epochs and the figure contains the average error in millimeters. Figure (a) shows the error as a function of the training epoch whereas figure (b) presents it as a function of training time. All MaskedFusion runs got smaller average error than DenseFusion}
    \label{fig:plot}
\end{figure}
In Fig. \ref{fig:plot} (a), we show the test results for several different epochs using the LineMOD dataset.
We trained both MaskedFusion and the DenseFusion for 100 epochs, and every 10 epochs we tested them and plotted their mean errors.
It can be seen in Fig. \ref{fig:plot} (a) that even our worst values are better than the best values of DenseFusion.
All MaskedFusion average error values were always below the DenseFusion in all epochs tested, and most important is that our method entered first in the $10mm$ error mark.
Our method achieves the $10mm$ error, on average, in epoch 30 and DenseFusion only achieved this error in epoch 40.
Comparing the mean error of our best run to the DenseFusion best result we have an error of $5.9mm$ and DenseFusion has $7.6mm$.
We achieved more accuracy leading to possible better placement of objects in a virtual scene or better grasping accuracy.
\begin{table}[thpb]
\centering
\caption{Quantitative evaluation of 6D pose using the ADD metric on the LineMOD dataset. Symmetric objects are presented in italic and were evaluated using ADD-S. Bold shows best results in a given row}
\label{table:compare}
\resizebox{\textwidth}{!}{%
\begin{tabular}{r|c|c|c|cc|ccccc}
\multicolumn{11}{c}{} \\
\multicolumn{4}{c}{}                                                                                            & \multicolumn{2}{c}{MaskedFusion} & \multicolumn{5}{c}{MaskedFusion}                          \\
\multicolumn{1}{c|}{Objects}   & SSD-6D + ICP & PointFusion & \multicolumn{1}{c|}{DenseFusion} & Avg   & Stdev & \multicolumn{5}{c}{Individual Experiments} \\ \hline
\multicolumn{1}{r|}{ape}       & 65.0         & 70.4        & \multicolumn{1}{c|}{92.3}        & 92.2  & 4.1   & \textbf{97.1}   & 86.7   & 90.5   & 91.4   & 95.2   \\
\multicolumn{1}{r|}{bench vi.} & 80.0         & 80.7        & \multicolumn{1}{c|}{93.2}        & 98.4  & 1.1   & 98.1   & \textbf{100.0}  & 97.1   & 99.0   & 98.1   \\
\multicolumn{1}{r|}{camera}    & 78.0         & 60.8        & \multicolumn{1}{c|}{94.4}        & 98.0  & 1.0   & 97.1   & 97.1   & \textbf{99.0}   & \textbf{99.0}   & 98.0   \\
\multicolumn{1}{r|}{can}       & 86.0         & 61.1        & \multicolumn{1}{c|}{93.1}        & 97.4  & 2.3   & 99.0   & 98.0   & 95.0   & \textbf{100.0}  & 95.0   \\
\multicolumn{1}{r|}{cat}       & 70.0         & 79.1        & \multicolumn{1}{c|}{96.5}        & 97.8  & 1.5   & 96.0   & 98.0   & 97.0   & \textbf{100.0}  & 98.0   \\
\multicolumn{1}{r|}{driller}   & 73.0         & 47.3        & \multicolumn{1}{c|}{87.0}        & 95.6  & 1.7   & 95.0   & 96.0   & 93.0   & \textbf{97.0}   & \textbf{97.0}   \\
\multicolumn{1}{r|}{duck}      & 66.0         & 63.0        & \multicolumn{1}{c|}{92.3}        & 94.0  & 3.0   & \textbf{97.2}   & 89.6   & 95.3   & 92.5   & 95.3   \\
\multicolumn{1}{r|}{\textit{eggbox}}    & \textbf{100.0}        & 99.9        & \multicolumn{1}{c|}{99.8}        & 99.6  & 0.5   & \textbf{100.0}  & \textbf{100.0}  & 99.1   & 99.1   & \textbf{100.0}  \\
\multicolumn{1}{r|}{\textit{glue}}      & \textbf{100.0}        & 99.3        & \multicolumn{1}{c|}{\textbf{100.0}}       & \textbf{100.0} & 0.0   & \textbf{100.0}  & \textbf{100.0}  & \textbf{100.0}  & \textbf{100.0}  & \textbf{100.0}  \\
\multicolumn{1}{r|}{hole p.}   & 49.0         & 71.8        & \multicolumn{1}{c|}{92.1}        & 97.3  & 2.5   & 93.3   & 97.1   & 98.1   & \textbf{100.0}  & 98.1   \\
\multicolumn{1}{r|}{iron}      & 78.0         & 83.2        & \multicolumn{1}{c|}{97.0}        & 97.1  & 1.3   & 95.9   & \textbf{99.0}   & 97.9   & 96.9   & 95.9   \\
\multicolumn{1}{r|}{lamp}      & 73.0         & 62.3        & \multicolumn{1}{c|}{95.3}        & 99.0  & 1.0   & 98.1   & \textbf{100.0}  & \textbf{100.0}  & 98.1   & 99.0   \\
\multicolumn{1}{r|}{phone}     & 79.0         & 78.8        & \multicolumn{1}{c|}{92.8}        & 98.8  & 1.3   & 97.1   & \textbf{100.0}  & \textbf{100.0}  & 99.0   & 98.1   \\ \hline
\multicolumn{1}{r|}{Average}   & 76.7         & 73.7        & \multicolumn{1}{c|}{94.3}        & 97.3  & 0.3   & 97.2   & 97.0   & 97.1   & \textbf{97.8}   & 97.5  
\end{tabular}%
}
\end{table}
To train our method for 100 epochs took 40 hours, compared with 33 for DenseFusion.
Since we have one more network to train and we have additional computation over the data and more data flowing in our method, it is normal to take longer in the overall training.
Note that, since we can achieve a smaller error before DenseFusion in terms of training epochs, this increase in training time can be disregarded, since, even if the training was stopped after a fixed number of hours instead of after a fixed number of epochs, MaskFusion would still produce a smaller estimation error, as can be seen in Fig. \ref{fig:plot} (b).
For instance, MaskedFusion entered in the $10mm$ error mark after 12 hours of training while DenseFusion needed 13.2 hours.

In Table \ref{table:compare} we present a comparison of our test results in a per object comparison with other three methods, SSD-6D \cite{ssd6d}, PointFusion \cite{pointfusion} and DenseFusion \cite{densefusion}.
The values presented in Table \ref{table:compare} result from the ADD metric (equation~\ref{eq:add}) and ADD-S metric (equation~\ref{eq:add-s}).
From Table \ref{table:compare}, we conclude that our method has overall better accuracy than previous methods.
\begin{table}[thpb]
\vspace{-2em}
\centering
\caption{Quantitative evaluation of 6D pose (area under the ADD-S (\ref{eq:add-s}) curve (AUC)) on the YCB-Video Dataset. Bold numbers are the best in a row and underline numbers are the best when comparing MaskedFusion with DenseFusion both with 100 training epochs. The last column of the table is the evaluation of MaskedFusion using the masks that were generated by our first sub-task during the train and test. (*) The values presented were obtained from \cite{densefusion}}
\label{table:AUC}
\resizebox{\textwidth}{!}{%
\begin{tabular}{r|c|c|c|cc||cc|cc||c}
\multicolumn{10}{c}{} \\
 & PointFusion & \begin{tabular}[c]{@{}c@{}}PoseCNN\\+ICP\end{tabular} & DenseFusion & \multicolumn{2}{c||}{MaskedFusion} & \multicolumn{2}{|c}{MaskedFusion} & \multicolumn{2}{|c||}{DenseFusion} & Pipeline  \\
 \multicolumn{1}{c|}{Training Epochs} & -- & -- & -- & \multicolumn{2}{c||}{200} & \multicolumn{2}{|c}{100} & \multicolumn{2}{|c||}{100} & 100
\\
\multicolumn{1}{c|}{Objects} & AUC(*) & AUC(*) & AUC(*) & \begin{tabular}[c]{@{}c@{}}AUC\\(Avg)\end{tabular} & \begin{tabular}[c]{@{}c@{}}AUC\\(Stdev)\end{tabular} & \begin{tabular}[c]{@{}c@{}}AUC\\(Avg)\end{tabular} & \begin{tabular}[c]{@{}c@{}}AUC\\(Stdev)\end{tabular}  & \begin{tabular}[c]{@{}c@{}}AUC\\(Avg)\end{tabular} & \begin{tabular}[c]{@{}c@{}}AUC\\(Stdev)\end{tabular} & AUC \\ \hline
002\_master\_chef\_can & 90.9 & 95.8 & \textbf{96.4} & 95.5 & 0.1 & \underline{95.9} & 0.1 & 94.3 & 0.7 & 95.0 \\
003\_cracker\_box & 80.5 & 92.7 & 95.5 & \textbf{96.7} & 0.4 & \underline{96.0} & 0.4 & 94.0 & 0.7 & 96.6 \\
004\_sugar\_box & 90.4 & \textbf{98.2} & 97.5 & 98.1 & 0.2 & \underline{97.6} & 0.2 & 95.7 & 1.6 & 98.2 \\
005\_tomato\_soup\_can & 91.9 & 94.5 & \textbf{94.6} & 94.3 & 0.1 & \underline{94.2} & 0.1 & 90.3 & 4.0 & 94.7 \\
006\_mustard\_bottle & 88.5 & \textbf{98.6} & 97.2 & 98.0 & 0.2 & \underline{97.6} & 0.2 & 95.2 & 1.8 & 98.0 \\
007\_tuna\_fish\_can & 93.8 & \textbf{97.1} & 96.6 & 96.9 & 0.3 & \underline{96.7} & 0.3 & 95.4 & 0.4 & 96.8 \\
008\_pudding\_box & 87.5 & \textbf{97.9} & 96.5 & 97.3 & 0.5 & \underline{96.3} & 0.5 & 95.1 & 0.8 & 98.2 \\
009\_gelatin\_box & 95.0 & \textbf{98.8} & 98.1 & 98.3 & 0.6 & \underline{98.0} & 0.6 & 97.2 & 0.5 & 98.8 \\
010\_potted\_meat\_can & 86.4 & \textbf{92.7} & 91.3 & 89.6 & 0.2 & \underline{89.4} & 0.2 & 88.1 & 0.5 & 91.0 \\
011\_banana & 84.7 & 97.1 & 96.6 & \textbf{97.6} & 0.1 & \underline{97.5} & 0.1 & 95.0 & 0.5 & 97.3 \\
019\_pitcher\_base & 85.5 & \textbf{97.8} & 97.1 & 97.7 & 0.3 & \underline{97.4} & 0.3 & 96.1 & 0.5 & 97.6 \\
021\_bleach\_cleanser & 81.0 & \textbf{96.9} & 95.8 & 95.4 & 0.8 & 93.8 & 0.8 & \underline{94.6} & 0.1 & 96.1 \\
024\_bowl & 75.7 & 81.0 & 88.2 & \textbf{89.6} & 3.1 & \underline{90.1} & 3.1 & 88.4 & 0.2 & 89.7 \\
025\_mug & 94.2 & 95.0 & \textbf{97.1} & \textbf{97.1} & 0.2 & \underline{97.0} & 0.2 & 95.8 & 0.5 & 97.2 \\
035\_power\_drill & 71.5 & \textbf{98.2} & 96.0 & 96.7 & 0.3 & \underline{96.4} & 0.3 & 93.9 & 1.1 & 96.6 \\
036\_wood\_block & 68.1 & 87.6 & 89.7 & \textbf{91.8} & 1.2 & \underline{90.6} & 1.2 & 90.2 & 0.8 & 90.7 \\
037\_scissors & 76.7 & 91.7 & \textbf{95.2} & 92.7 & 0.6 & \underline{93.2} & 0.6 & 92.4 & 1.3 & 90.6 \\
040\_large\_marker & 87.9 & 97.2 & \textbf{97.5} & \textbf{97.5} & 0.3 & \underline{97.0} & 0.3 & 95.7 & 0.5 & 97.1 \\
051\_large\_clamp & 65.9 & \textbf{75.2} & 72.9 & 71.9 & 1.3 & \underline{72.1} & 1.3 & 69.7 & 0.4 & 74.7 \\
052\_extra\_large\_clamp & 60.4 & 64.4 & 69.8 & \textbf{71.4} & 1.0 & \underline{69.6} & 1.0 & 64.5 & 0.6 & 58.8 \\
061\_foam\_brick & 91.8 & \textbf{97.2} & 92.5 & 94.3 & 1.1 & \underline{94.2} & 1.1 & 92.0 & 2.0 & 93.7 \\ \hline 
Average & 83.9 & 93.0 & 93.1 & \textbf{93.3} & 0.6 & \underline{92.9} & 0.6 & 91.1 & 0.9 & 92.7
\end{tabular}%
}
\end{table}
The average results from the 5 repetitions of our method are better than DenseFusion in 11 out of 13 objects.
In our worst-performing experience, the second column of MaskedFusion Individual Experiments in Table \ref{table:compare}, we achieved an average of $97\%$, which is better than the DenseFusion.
Finally, the best of our 5 repetitions improves the overall ADD from the $94.3\%$ of DenseFusion to $97.8\%$.
For the LineMOD dataset, we have achieved overall better results than previous methods.

\subsection{Results in YCB-Video}
We also use the YCB-Video dataset to evaluate our method and compare it with other methods.
The results presented in Table \ref{table:AUC} show the area under the accuracy-threshold curve (AUC) using the ADD-S metric with the maximum threshold of $10cm$.

As shown in Table \ref{table:AUC}, MaskedFusion obtained the best average score using 200 epochs for training.
Since DenseFusion, that was the closest one to us, didn't state how many epochs they used, we also present a comparison between MaskedFusion and DenseFusion using 100 epochs for training.
All of our experiments have 5 repetitions so we can compute the average of all executions, this also includes the experiment that we have done on DenseFusion with 100 training epochs.
MaskedFusion 6D pose NN achieved an average of more $0.2\%$ in the AUC than what DenseFusion presented.
But when we compare the average of our method using 100 training epochs and DenseFusion we achieved an an improvement of $1.8\%$ AUC score.

On Table \ref{table:2cm} we present the percentage of ADD-S smaller than $2cm$.
This metric was introduced in \cite{densefusion} since predictions under $2cm$ are the minimum tolerance for robot manipulation.

As in the AUC, DenseFusion did not state how many epochs were used for training, so for this metric, we also did the same as in AUC.
We trained MaskedFusion and DenseFusion 100 epochs to compare their $<2cm$ metric and we had an average improvement of $0.3\%$.
\begin{table}[thpb]
\vspace{-2em}
\centering
\caption{Quantitative evaluation of 6D pose (percentage of ADD-S smaller than $2cm$) on the YCB-Video Dataset. Bold numbers are the best in a row and underline numbers are the best when comparing MaskedFusion with DenseFusion both with 100 training epochs. The last column of the table is the evaluation of MaskedFusion using the masks that were generated by our first sub-task during the train and test. (*) The values presented were obtained from \cite{densefusion}}
\label{table:2cm}
\resizebox{\textwidth}{!}{%
\begin{tabular}{r|c|c|c|cc||cc|cc||c}
\multicolumn{10}{c}{} \\
 & PointFusion & \begin{tabular}[c]{@{}c@{}}PoseCNN\\+ICP\end{tabular} & DenseFusion & \multicolumn{2}{c||}{MaskedFusion} & \multicolumn{2}{|c}{MaskedFusion} & \multicolumn{2}{|c||}{DenseFusion} & Pipeline \\
  \multicolumn{1}{c|}{Training Epochs} & -- & -- & -- & \multicolumn{2}{c||}{200} & \multicolumn{2}{|c}{100} & \multicolumn{2}{|c||}{100} & 100
\\
\multicolumn{1}{c|}{Objects} & \textless{}2cm(*) & \textless{}2cm(*) & \textless{}2cm(*) & \begin{tabular}[c]{@{}c@{}}\textless{}2cm\\(Avg)\end{tabular} & \begin{tabular}[c]{@{}c@{}}\textless{}2cm\\(Stdev)\end{tabular} & \begin{tabular}[c]{@{}c@{}}\textless{}2cm\\(Avg)\end{tabular} & \begin{tabular}[c]{@{}c@{}}\textless{}2cm\\(Stdev)\end{tabular} & \begin{tabular}[c]{@{}c@{}}\textless{}2cm\\(Avg)\end{tabular} & \begin{tabular}[c]{@{}c@{}}\textless{}2cm\\(Stdev)\end{tabular} & \textless{}2cm \\ \hline
002\_master\_chef\_can & 99.8 & \textbf{100.0} & \textbf{100.0} & \textbf{100.0} & 0.0 & \underline{100.0} & 0.0 & \underline{100.0} & 0.0 & 100.0 \\
003\_cracker\_box & 62.6 & 91.6 & 99.5 & \textbf{99.8} & 0.1 & \underline{99.7} & 0.1 & 99.5 & 0.5 & 99.3 \\
004\_sugar\_box & 95.4 & \textbf{100.0} & \textbf{100.0} & \textbf{100.0} & 0.0 & \underline{100.0} & 0.0 & \underline{100.0} & 0.0 & 100.0 \\
005\_tomato\_soup\_can & \textbf{96.9} & \textbf{96.9} & \textbf{96.9} & \textbf{96.9} & 0.0 & \underline{96.5} & 0.0 & 92.6 & 5.9 & 96.9 \\
006\_mustard\_bottle & 84.0 & \textbf{100.0} & \textbf{100.0} & \textbf{100.0} & 0.0 & \underline{100.0} & 0.0 & 99.9 & 0.1 & 100.0 \\
007\_tuna\_fish\_can & 99.8 & \textbf{100.0} & \textbf{100.0} & 99.7 & 0.1 & 99.9 & 0.1 & \underline{100.0} & 0.0 & 100.0 \\ 
008\_pudding\_box & 96.7 & \textbf{100.0} & \textbf{100.0} & \textbf{100.0} & 0.0 & \underline{100.0} & 0.0 & \underline{100.0} & 0.0 & 100.0 \\
009\_gelatin\_box & \textbf{100.0} & \textbf{100.0} & \textbf{100.0} & \textbf{100.0} & 0.0 & \underline{100.0} & 0.0 & \underline{100.0} & 0.0 & 100.0 \\ 
010\_potted\_meat\_can & 88.5 & 93.6 & 93.1 & \textbf{94.2} & 0.4 & \underline{90.8} & 0.4 & 90.6 & 0.2 & 92.8 \\
011\_banana & 70.5 & 99.7 & \textbf{100.0} & \textbf{100.0} & 0.0 & \underline{100.0} & 0.0 & 99.7 & 0.5 & 99.7 \\
019\_pitcher\_base & 79.8 & \textbf{100.0} & \textbf{100.0} & \textbf{100.0} & 0.3 & 99.9 & 0.3 & \underline{100.0} & 0.0 & 99.8 \\
021\_bleach\_cleanser & 65.0 & 99.4 & \textbf{100.0} & 99.4 & 3.9 & 94.0 & 3.9 & \underline{100.0} & 0.0 & 99.6 \\
024\_bowl & 24.1 & 54.9 & \textbf{98.8} & 95.4 & 5.8 & \underline{95.8} & 5.8 & 95.0 & 4.2 & 91.6 \\
025\_mug & 99.8 & 99.8 & \textbf{100.0} & \textbf{100.0} & 0.0 & \underline{100.0} & 0.0 & \underline{100.0} & 0.0 & 100.0 \\
035\_power\_drill & 22.8 & \textbf{99.6} & 98.7 & 99.5 & 0.3 & \underline{99.4} & 0.3 & 97.1 & 1.5 & 99.6 \\
036\_wood\_block & 18.2 & 80.2 & 94.6 & \textbf{100.0} & 0.9 & \underline{98.1} & 0.9 & 97.5 & 1.5 & 90.8 \\
037\_scissors & 35.9 & 95.6 & \textbf{100.0} & 99.9 & 1.1 & 99.0 & 1.1 & \underline{99.4} & 0.7 & 92.8 \\
040\_large\_marker & 80.4 & 99.7 & \textbf{100.0} & 99.9 & 0.1 & 99.7 & 0.1 & \underline{99.8} & 0.1 & 99.4 \\
051\_large\_clamp & 50.0 & 74.9 & \textbf{79.2} & 78.7 & 0.1 & \underline{77.9} & 0.1 & 75.6 & 1.2 & 80.8 \\
052\_extra\_large\_clamp & 20.1 & 48.8 & \textbf{76.3} & 75.9 & 2.1 & \underline{72.0} & 2.1 & 68.7 & 0.7 & 75.3 \\
061\_foam\_brick & \textbf{100.0} & \textbf{100.0} & \textbf{100.0} & \textbf{100.0} & 0.0 & \underline{100.0} & 0.0 & \underline{100.0} & 0.0 & 99.3 \\ \hline 
Average & 74.1 & 93.2 & 96.8 & \textbf{97.1} & 0.7 & \underline{96.3} & 0.7 & 96.0 & 0.8 & 96.1
\end{tabular}%
}
\end{table}

\begin{figure}[thpb]
\vspace{-2em}
    \centering
    \subfloat[Error per epoch]{{\includegraphics[width=.45\linewidth]{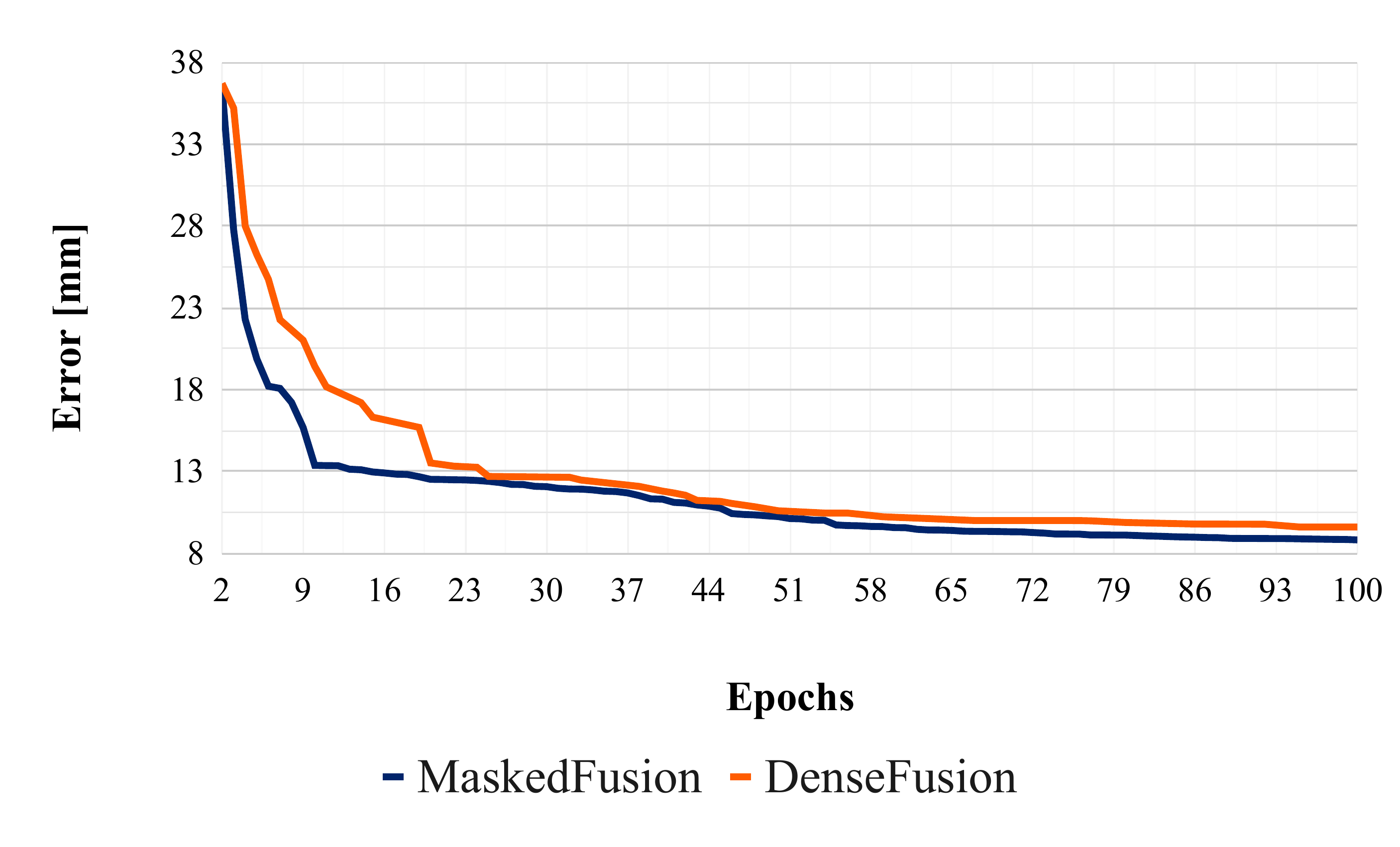} }}%
    \qquad
    \subfloat[Error per hour]{{\includegraphics[width=.45\linewidth]{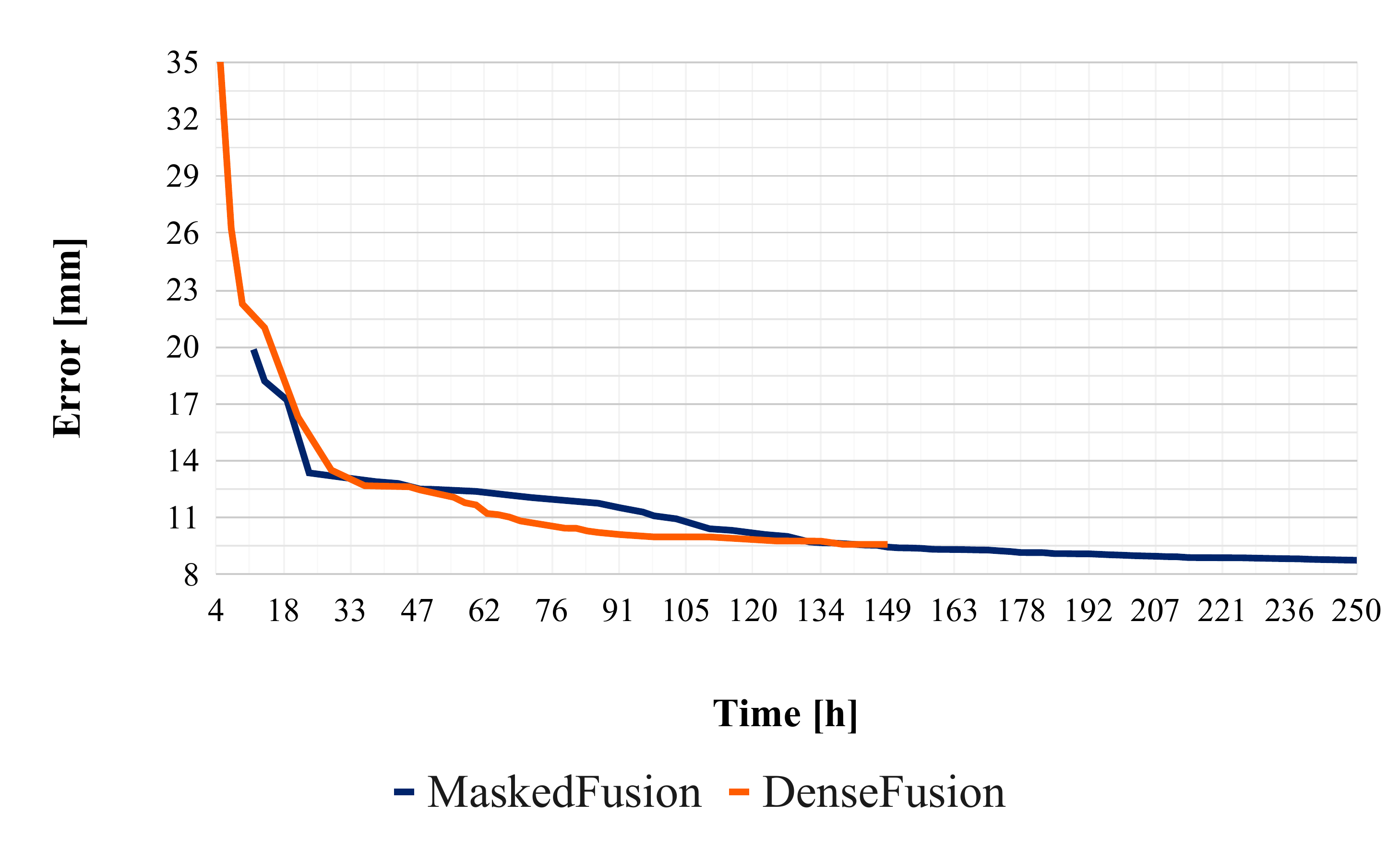} }}%
\subsection{Results in YCB-Video}
    \label{fig:plotycb}
\end{figure}

As in LineMOD we have one more network to train and we have additional computation over the data and more data flowing in our method, so we took more 95 of hours to train 100 epochs.
MaskedFusion took 240 hours compared with 145 hours of training time for DenseFusion.
Fig. \ref{fig:plotycb} presents these results both in terms of epochs and in terms of time.

\subsection{Pipeline results in inference time}
\label{sec:inference}
The inference time of MaskedFusion in the first sub-task (image segmentation) is $0.2$ seconds per image.
For the second sub-task (6D pose estimation) $0.01$ seconds were needed to estimate the pose per given object, and on the third sub-task, the pose refinement took $0.002$ seconds.
So the total inference time of MaskedFusion is $0.212$ seconds from the moment of the RGB-D image capture until it has the estimated pose.

Using the masks that were generated by our first sub-task during the train we achieved $92.7\%$ in the AUC metrics (shown on the last column of Table \ref{table:AUC}) and $96.1\%$ in the $<2cm$ metric (shown on the last column of Table \ref{table:2cm}).

Using our masks we had worst overall scores as expected, since we were not using corrected
masks as those provided with YCB.
With this test we have learned that our semantic segmentation needs to be improved to obtain better masks so that our second sub-task can achieve better 6D pose estimation.
We note that other papers in the literature do not present the results on the full pipeline and always consider that the method receives perfectly segmented data.

\section{Conclusion}
Achieving a robust estimation of the 6D pose of objects captured in the real world is still an open challenge.
MaskedFusion improved the state-of-the-art in this area using objects' masks retrieved during the semantic segmentation, to identify and localize the object in the RGB image.
We achieved an error bellow $6mm$ in LineMOD with just 100 training epochs and in YCB-Video, a challenging dataset, we have obtained AUC score of $93.3\%$ and $97.1\%$ in the \textless{}2cm metric.
The masks are used to remove non-relevant data from the input of the NN and serve as an additional feature to the 6D pose estimation NN.
MaskedFusion has low inference time but at the cost of an increase in training time.
We saw that this increased training time can sometimes (such as with LineMOD data and during most of YCB-Video training) be disregarded since we still beat the state-of-the-art if the training time stops after a fixed number of hours instead of a fixed number of epochs.
As future work, we intend to study the influence of the instance segmentation method on the MaskedFusion results (since the higher the accuracy of the first sub-task, the better results are expected from the second sub-task) and work towards speeding up the training stage of MaskedFusion.


%
%
\bibliographystyle{splncs04}
\bibliography{biblio}

\begin{thebibliography}{10}
\providecommand{\url}[1]{\texttt{#1}}
\providecommand{\urlprefix}{URL }
\providecommand{\doi}[1]{https://doi.org/#1}

\bibitem{segnet}
Badrinarayanan, V., Kendall, A., Cipolla, R.: Segnet: A deep convolutional
  encoder-decoder architecture for image segmentation. IEEE transactions on
  pattern analysis and machine intelligence  \textbf{39}(12),  2481--2495
  (2017)

\bibitem{pnpex1}
Brachmann, E., Krull, A., Michel, F., Gumhold, S., Shotton, J., Rother, C.:
  Learning 6d object pose estimation using 3d object coordinates. In: European
  conference on computer vision. pp. 536--551. Springer (2014)

\bibitem{lineex1}
Buch, A.G., Kiforenko, L., Kraft, D.: Rotational subgroup voting and pose
  clustering for robust 3d object recognition. In: 2017 IEEE International
  Conference on Computer Vision (ICCV). pp. 4137--4145. IEEE (2017)

\bibitem{originalycb}
Calli, B., Singh, A., Walsman, A., Srinivasa, S., Abbeel, P., Dollar, A.M.: The
  ycb object and model set: Towards common benchmarks for manipulation
  research. In: 2015 international conference on advanced robotics (ICAR). pp.
  510--517. IEEE (2015)

\bibitem{lineex2}
Drost, B., Ulrich, M., Navab, N., Ilic, S.: Model globally, match locally:
  Efficient and robust 3d object recognition. In: 2010 IEEE computer society
  conference on computer vision and pattern recognition. pp. 998--1005. IEEE
  (2010)

\bibitem{pnp}
Fischler, M.A., Bolles, R.C.: Random sample consensus: a paradigm for model
  fitting with applications to image analysis and automated cartography.
  Communications of the ACM  \textbf{24}(6),  381--395 (1981)

\bibitem{linemod}
Hinterstoisser, S., Holzer, S., Cagniart, C., Ilic, S., Konolige, K., Navab,
  N., Lepetit, V.: Multimodal templates for real-time detection of texture-less
  objects in heavily cluttered scenes. In: 2011 international conference on
  computer vision. pp. 858--865. IEEE (2011)

\bibitem{ssd6d}
Kehl, W., Manhardt, F., Tombari, F., Ilic, S., Navab, N.: Ssd-6d: Making
  rgb-based 3d detection and 6d pose estimation great again. In: Proceedings of
  the IEEE International Conference on Computer Vision. pp. 1521--1529 (2017)

\bibitem{localrgbd}
Kehl, W., Milletari, F., Tombari, F., Ilic, S., Navab, N.: Deep learning of
  local rgb-d patches for 3d object detection and 6d pose estimation. In:
  European Conference on Computer Vision. pp. 205--220. Springer (2016)

\bibitem{panopticsegmentation}
Kirillov, A., He, K., Girshick, R., Rother, C., Doll{\'a}r, P.: Panoptic
  segmentation. In: Proceedings of the IEEE conference on computer vision and
  pattern recognition. pp. 9404--9413 (2019)

\bibitem{VGG16}
Krizhevsky, A., Sutskever, I., Hinton, G.E.: Imagenet classification with deep
  convolutional neural networks. In: Advances in neural information processing
  systems. pp. 1097--1105 (2012)

\bibitem{liunified}
Li, C., Bai, J., Hager, G.D.: A unified framework for multi-view multi-class
  object pose estimation. In: Proceedings of the European Conference on
  Computer Vision (ECCV). pp. 254--269 (2018)

\bibitem{pvfh}
Li, D., Liu, N., Guo, Y., Wang, X., Xu, J.: 3d object recognition and pose
  estimation for random bin-picking using partition viewpoint feature
  histograms. Pattern Recognition Letters  \textbf{128},  148--154 (2019)

\bibitem{lineex4}
Li, Y., Wang, G., Ji, X., Xiang, Y., Fox, D.: Deepim: Deep iterative matching
  for 6d pose estimation. In: Proceedings of the European Conference on
  Computer Vision (ECCV). pp. 683--698 (2018)

\bibitem{deepex2}
Mousavian, A., Anguelov, D., Flynn, J., Kosecka, J.: 3d bounding box estimation
  using deep learning and geometry. In: Proceedings of the IEEE Conference on
  Computer Vision and Pattern Recognition. pp. 7074--7082 (2017)

\bibitem{pnpex2}
Pavlakos, G., Zhou, X., Chan, A., Derpanis, K.G., Daniilidis, K.: 6-dof object
  pose from semantic keypoints. In: 2017 IEEE International Conference on
  Robotics and Automation (ICRA). pp. 2011--2018. IEEE (2017)

\bibitem{pvnet}
Peng, S., Liu, Y., Huang, Q., Zhou, X., Bao, H.: Pvnet: Pixel-wise voting
  network for 6dof pose estimation. In: Proceedings of the IEEE Conference on
  Computer Vision and Pattern Recognition. pp. 4561--4570 (2019)

\bibitem{frustum}
Qi, C.R., Liu, W., Wu, C., Su, H., Guibas, L.J.: Frustum pointnets for 3d
  object detection from rgb-d data. In: Proceedings of the IEEE Conference on
  Computer Vision and Pattern Recognition. pp. 918--927 (2018)

\bibitem{pointnet}
Qi, C.R., Su, H., Mo, K., Guibas, L.J.: Pointnet: Deep learning on point sets
  for 3d classification and segmentation. In: Proceedings of the IEEE
  Conference on Computer Vision and Pattern Recognition. pp. 652--660 (2017)

\bibitem{unet}
Ronneberger, O., Fischer, P., Brox, T.: U-net: Convolutional networks for
  biomedical image segmentation. In: International Conference on Medical image
  computing and computer-assisted intervention. pp. 234--241. Springer (2015)

\bibitem{fpfh}
Rusu, R.B.: Semantic 3D Object Maps for Everyday Manipulation in Human Living
  Environments. Ph.D. thesis, Computer Science department, Technische
  Universitaet Muenchen, Germany (October 2009)

\bibitem{vfh}
Rusu, R.B., Bradski, G., Thibaux, R., Hsu, J.: Fast 3d recognition and pose
  using the viewpoint feature histogram. In: Proceedings of the 23rd IEEE/RSJ
  International Conference on Intelligent Robots and Systems (IROS). Taipei,
  Taiwan (October 2010)

\bibitem{deepex1}
Suwajanakorn, S., Snavely, N., Tompson, J.J., Norouzi, M.: Discovery of latent
  3d keypoints via end-to-end geometric reasoning. In: Advances in Neural
  Information Processing Systems. pp. 2059--2070 (2018)

\bibitem{tejani}
{Tejani}, A., {Kouskouridas}, R., {Doumanoglou}, A., {Tang}, D., {Kim}, T.:
  Latent-class hough forests for 6 dof object pose estimation. IEEE
  Transactions on Pattern Analysis and Machine Intelligence  \textbf{40}(1),
  119--132 (Jan 2018). \doi{10.1109/TPAMI.2017.2665623}

\bibitem{pnpex3}
Tekin, B., Sinha, S.N., Fua, P.: Real-time seamless single shot 6d object pose
  prediction. In: Proceedings of the IEEE Conference on Computer Vision and
  Pattern Recognition. pp. 292--301 (2018)

\bibitem{pnpex4}
Tremblay, J., To, T., Sundaralingam, B., Xiang, Y., Fox, D., Birchfield, S.:
  Deep object pose estimation for semantic robotic grasping of household
  objects. In: Conference on Robot Learning. pp. 306--316 (2018)

\bibitem{lineex3}
Vidal, J., Lin, C.Y., Mart{\'\i}, R.: 6d pose estimation using an improved
  method based on point pair features. In: 2018 4th International Conference on
  Control, Automation and Robotics (ICCAR). pp. 405--409. IEEE (2018)

\bibitem{densefusion}
Wang, C., Xu, D., Zhu, Y., Mart{\'\i}n-Mart{\'\i}n, R., Lu, C., Fei-Fei, L.,
  Savarese, S.: Densefusion: 6d object pose estimation by iterative dense
  fusion. In: Proceedings of the IEEE Conference on Computer Vision and Pattern
  Recognition. pp. 3343--3352 (2019)

\bibitem{wu2018real}
Wu, J., Zhou, B., Russell, R., Kee, V., Wagner, S., Hebert, M., Torralba, A.,
  Johnson, D.M.: Real-time object pose estimation with pose interpreter
  networks. In: 2018 IEEE/RSJ International Conference on Intelligent Robots
  and Systems (IROS). pp. 6798--6805. IEEE (2018)

\bibitem{posecnn}
Xiang, Y., Schmidt, T., Narayanan, V., Fox, D.: Posecnn: A convolutional neural
  network for 6d object pose estimation in cluttered scenes. arXiv preprint
  arXiv:1711.00199  (2017)

\bibitem{pointfusion}
Xu, D., Anguelov, D., Jain, A.: Pointfusion: Deep sensor fusion for 3d bounding
  box estimation. In: Proceedings of the IEEE Conference on Computer Vision and
  Pattern Recognition. pp. 244--253 (2018)

\bibitem{pspnet}
Zhao, H., Shi, J., Qi, X., Wang, X., Jia, J.: Pyramid scene parsing network.
  In: Proceedings of the IEEE conference on computer vision and pattern
  recognition. pp. 2881--2890 (2017)

\bibitem{voxelnet}
Zhou, Y., Tuzel, O.: Voxelnet: End-to-end learning for point cloud based 3d
  object detection. In: Proceedings of the IEEE Conference on Computer Vision
  and Pattern Recognition. pp. 4490--4499 (2018)

\end{thebibliography}
\end{document}